\documentclass{article}

\usepackage{PRIMEarxiv}
\usepackage[utf8]{inputenc}
\usepackage[T1]{fontenc}
\usepackage{hyperref}
\usepackage{url}
\usepackage{booktabs}
\usepackage{amsmath}
\usepackage{amsfonts}
\usepackage{microtype}
\usepackage{graphicx}
\usepackage{multirow}
\usepackage{array}
\usepackage{tabularx}
\usepackage{makecell}
\graphicspath{{figures/}}


\newcommand{\method}{MKGR}
\newcommand{\bce}{\ell_{\mathrm{bce}}}

\title{MKGR: Multimodal Knowledge-Graph Representation Learning for Cold-Start Protein-Protein Interaction Prediction}

\author{
  Wenbo Zhang \\
  College of Computer and Information Science, Southwest University, Chongqing \\
  Chongqing, China \\
  \texttt{z13996091260@email.swu.edu.cn}
}

\begin{document}
\maketitle

\begin{abstract}
Accurate protein-protein interaction (PPI) prediction is central to functional genomics, disease mechanism discovery, and drug development. A difficult setting arises when candidate interactions include proteins that have no observed PPI edges during training, where models relying on network topology alone often lose useful context. This paper presents \method, a multimodal representation framework for cold-start PPI prediction. \method\ combines region-aware protein sequence encoding with four protein-centered biomedical knowledge graphs, including protein-drug, protein-disease, protein-miRNA, and protein-lncRNA associations. The sequence branch extracts contextual representations from structurally informed sequence regions, while graph attention encoders learn modality-specific protein embeddings from sparse biomedical associations. A bridge reconstruction objective regularizes graph learning by recovering shared protein-entity associations, and a pair-level gating module adaptively integrates sequence and graph evidence for each candidate protein pair. Experiments on two benchmark datasets under novel-old and novel-novel cold-start settings show that \method\ consistently outperforms competitive sequence, network, and knowledge-graph baselines across ACC, F1, AUC, AUPR, and MCC.
\end{abstract}


\section{Introduction}
Protein-protein interactions organize many cellular processes, including signal transduction, transcriptional regulation, immune response, and metabolic control. Experimental assays such as yeast two-hybrid screening and affinity purification followed by mass spectrometry have produced large interaction resources, but they remain costly, condition-dependent, and incomplete. Computational PPI prediction therefore plays an important role in prioritizing candidate interactions and expanding biological networks beyond experimentally verified edges.

Recent deep learning models have improved PPI prediction by encoding amino-acid sequences, protein structures, and interaction networks. Sequence-based models benefit from pretrained protein language models and convolutional or attention-based encoders, while graph models exploit relational structure among proteins or between proteins and biomedical entities. However, the cold-start case remains challenging: for a newly introduced protein, observed PPI edges are unavailable or sparse, so a predictor must transfer information from intrinsic sequence patterns and external biomedical associations. Relying on only one evidence source is often fragile. Sequence encoders may miss disease-, drug-, or RNA-related context, whereas graph encoders may suffer when a protein has limited coverage in a particular biomedical modality.

This paper studies cold-start PPI prediction with a multimodal sequence-graph design. We propose \method, a representation learning framework that models both intrinsic protein sequence signals and extrinsic biomedical graph signals. In the sequence branch, each protein is divided into structural regions and encoded with a pretrained protein language model followed by a Transformer encoder. In the graph branch, four modality-specific protein-entity graphs are modeled with graph attention networks. To make sparse graph learning more robust, \method\ introduces a bridge reconstruction task that encourages protein and entity embeddings to preserve shared biomedical associations. Finally, pair-level gated fusion combines sequence and graph representations in a candidate-specific manner, allowing the model to adjust the contribution of each modality for each protein pair.

The main contributions are as follows. First, we formulate a cold-start PPI model that jointly uses protein sequences and multimodal biomedical knowledge graphs. Second, we design a bridge reconstruction objective for graph-regularized protein representation learning under sparse protein-entity associations. Third, we introduce pair-level gated fusion to adaptively integrate heterogeneous modalities. Fourth, experiments on two datasets and two cold-start tasks demonstrate consistent gains over representative baselines.

\section{Related Work}
\textbf{Computational PPI prediction.}
Early computational PPI methods used protein descriptors, kernels, matrix factorization, or network propagation to infer missing interactions from biological features and known interaction networks. Recent deep models have shifted toward end-to-end representation learning over sequences, structures, and interaction neighborhoods. TAGPPI integrates sequence features with structural contact maps, while HNSPPI combines sequence and network evidence for supervised PPI prediction \cite{TAGPPI,HNSPPI}. EResCNN uses ensemble residual convolution over sequence descriptors, and BaPPI studies balanced learning for interaction classification \cite{EResCNN,BaPPI}. Sequence-based PPI predictors such as PIPR and D-SCRIPT further show that protein sequences can support interaction inference when structure or network information is incomplete \cite{pipr2019,dscript2021}. Protein resources and structure predictors, including UniProt, ESM, ESMFold, and AlphaFold, have also strengthened sequence-centered protein representation learning \cite{uniprot2021,esm1b,esmfold,alphafold2021}. These methods provide powerful protein representations, but they still face difficulty when test proteins have no observed PPI links during training.

\textbf{Protein sequence and biomedical graph evidence.}
Sequence-only models capture intrinsic amino-acid patterns, whereas graph-based models can exploit external biomedical associations that are not directly available from the PPI network. Graph neural networks provide a general mechanism for neighborhood aggregation and inductive node representation learning \cite{gcn,graphsage}. Attention-based and expressive message-passing variants further improve local relation modeling in heterogeneous or sparse graphs \cite{gat,gin}, and graph benchmarks support systematic evaluation of such methods \cite{ogb2020}. In biomedical settings, resources such as STRING, DrugBank, CTD, miRTarBase, and LncTarD connect proteins to interaction partners, chemicals, diseases, miRNAs, and lncRNAs \cite{string,drugbank,ctd,mirtarbase,lnctard}. Interaction and ontology resources such as IntAct and the Gene Ontology provide additional curated biological context \cite{intact2014,geneontology2021}. Larger biomedical knowledge graphs such as Hetionet and PrimeKG show that integrating heterogeneous biomedical entities can support downstream discovery tasks \cite{hetionet,primekg}. Multi-relational embedding and relational GCN models provide general tools for knowledge-graph representation \cite{transe2013,rgcn2018}, and graph models have also been used for biomedical relation prediction such as polypharmacy side-effect modeling \cite{decagon2018}. Knowledge-graph-enhanced PPI models, including KGF-GNN and HEENN, demonstrate the value of protein-associated graph context \cite{KGF,HEENN}. Multimodal knowledge graph fusion has also been used for related biomedical relation prediction problems such as drug-drug interaction prediction \cite{MKGFENN}.

\textbf{Sparse and incomplete-data representation.}
Cold-start PPI prediction can be viewed as a sparse representation problem in which different proteins have uneven information coverage across modalities. Latent factor models and graph-regularized factorization methods have long studied how to learn compact representations from sparse and incomplete observations \cite{xlYu2026FederatLatentFactorLearnin1,WuTKDE2022,xlChen2023ADiffereEvolutiEnhance4,xlXu2025RecursiandFuzzineReinfor2,xlLiang2023MMAMultiMetricAutoenc5,xlXu2025AHighlyAccuratThree3}. Adaptive regularization, non-negative latent factor modeling, and dynamic sparse tensor estimation further improve robustness when data are high-dimensional or partially observed \cite{xlLuo2020AdaptivRegularIncorpoLatent9,xlLuo2021AlgoritofUnconstNon8,xlYuan2020AGeneralandFast10,xlLi2026NeuralNonnegaLatentFactori32,xlXu2026ASamplinNeighboRegular29}. Federated latent factor learning, differential-evolution-enhanced latent factor analysis, multi-metric autoencoders, and adaptive regularization all show how auxiliary optimization or structural constraints can be integrated into representation learning.

\textbf{Latent factor learning and streaming features.}
Several studies by Luo and collaborators investigate sparse matrix representation, service QoS prediction, and non-negative latent factor learning from complementary viewpoints. Prediction-sampling and double-space modeling provide two additional examples of robust latent representation under sparse observations \cite{xlXu2023OnlineSparseStreamiFeature7,wuPNMLF2022,xlZhang2023AnErrorCorrectMid6,wuTSC2023}. Online feature selection is another relevant line because cold-start biological data often arrive with changing or incomplete feature sets. Tensor compression and related low-rank representation studies also support robust representation learning for sparse or incomplete observations \cite{xlHe2026Adaptivtuckerdecompobased11,xlHe2026TensorLowRankOrthogo14}. Time-dependent incomplete data studies further motivate designs that adapt to changing observations.

\textbf{High-order, tensor, and graph representation.}
Auxiliary relational structure is often useful when target labels are sparse. Tensor and high-order graph models are closely related to multimodal biomedical learning because multiple entity types naturally induce high-order relations. Adaptive Tucker decomposition, tensor low-rank compression, and neural Tucker factorization provide representative examples of high-order representation learning \cite{xlHe2026Adaptivtuckerdecompobased11,xlHe2026TensorLowRankOrthogo14,xlTang2025NeuralTuckerFactori37}. Multi-aspect self-attending neural Tucker factorization and multi-projection self-attending Tucker models further extend tensor learning to spatiotemporal and incomplete data \cite{xlHou2026MultiAspectSelfAttendi17,xlTang2026MPSANTAnovelmulti19}. Graph convolutional and modular graph models provide additional strategies for propagating information over structured data \cite{xlLin2026DualChannelGraphConvolu13,xlHe2026ModularGraphConvoluNetwork16}. Graph-incorporated factor analysis and multi-metric latent feature analysis also connect sparse representation learning with graph or metric structure \cite{xlWang2026AdvanceHighOrderGraph15,wuGraphLFA2023,xlWang2026GraphTensorConvoluNetwork33,wuMMLF2024}. Broader surveys and tensor-based causal or spatiotemporal models show the continuing development of high-order representation learning \cite{xlHe2026Asurveyoflatent18,xlLiao2025ANovelTensorCausal48,xlChen2025LatentFactoriofTensors38}.

\textbf{Recent multimodal applications.}
Recent representation-learning studies have explored cross-modal medical learning, semantic segmentation distillation, scalable sentiment analysis, mixture-of-experts learning, and neural community search \cite{xlLan2026CMCGNSCrossmodal12,xlGou2026Layerwisecorrelaand20,xlLiu2026AScalablMultichSentime21,xlDeng2026FuzzyMixtureofExperts22,xlLin2026NCSACEffectiNeuralCommuni27}. Other works study robust optimization, contrastive distillation for electricity-theft detection, knowledge-driven multiple-instance learning, dynamic optimization, and precise latent factor analysis \cite{xlYang2026AnIntelliOptimizBased23,xlQin2026ARobustApproacto24,xlLi2026KnowledDrivenMultiplInstanc25,xlLyu2026DynamicStochasReorienParticl26,xlLyu2026GeneticAlgoritBasedTwo28}. Contrastive and collaborative distillation approaches have also been used for medical, graph-structured, and service data representation \cite{xlGou2026MultiScaleCollaboDistill30,xlHan2026TraceHGAnUnsuperDual31,xlWei2025CLORGAcontraslearnin34,xlQiao2025Identifnoveltherapetargets35}. Industrial and educational applications, including distributed optimization, student emotion analysis, PID-enhanced factor analysis, robot calibration, and traffic-data imputation, illustrate the breadth of multimodal representation learning under heterogeneous observations \cite{xlLiao2025LocalSearchBasedAnytime36,xlLuo2025AnalysiofStudentPositiv39,xlYuan2025AProportIntegraControl40,xlChen2025DataDrivenCalibraof41,xlLin2025ADConvoluIncorpo50,xlYang2025LatentFactorAnalysiModel51}. Biomedical and healthcare-oriented studies, including drug repositioning, medical image representation, large language models for healthcare, organoid classification, and larynx pathological grading, are particularly close to the multimodal biomedical motivation of this work \cite{xlZhao2025Regulatawaregraphlearnin46,xlHuang2025FDTsAFeatureDisenta42,xlHu2025AdvanciHealthcwithLarge43,xlWei2025CLORGAcontraslearnin34,xlLi2026KnowledDrivenMultiplInstanc25}. Additional studies on matrix factorization review, deep latent factor learning, graph convolution enhancement, robot calibration, posterior-neighborhood regularization, traffic prediction, learning-error refinement, and hash-factor representation show that robust representation learning remains broadly useful across domains \cite{xlHu2025AComprehReviewof44,wuDeepLFM2021,xlChen2025Enhancigraphconvolunetwork45,xlLi2025SearchiforanAccurat47,wuPosterior2022,xlLi2025LearninErrorRefinemin49,xlChen2025DataDrivenCalibraof41,wuTPAMI2026}. \method\ follows this broad representation-learning view but specializes it for cold-start PPI prediction by combining modality-specific graph encoders, bridge reconstruction, and adaptive fusion across sequence and biomedical graph branches.

\section{Preliminaries}
\textbf{Cold-start PPI setting.}
Let $\mathcal{P}$ be the protein universe and let $\Omega \subseteq \mathcal{P}\times\mathcal{P}$ be the set of candidate pairs with binary labels $y_{ij}\in\{0,1\}$. Proteins are split into an observed subset $\mathcal{P}_{o}$ and a disjoint novel subset $\mathcal{P}_{n}$. The learner is fitted only on labeled pairs from $\mathcal{P}_{o}\times\mathcal{P}_{o}$ and is evaluated on two transfer regimes: $\mathcal{P}_{n}\times\mathcal{P}_{o}$ and $\mathcal{P}_{n}\times\mathcal{P}_{n}$. Thus, the prediction function must infer an interaction probability
\begin{equation}
  \hat{y}_{ij}=F_{\Theta}(x_i,x_j,\mathcal{G}_i,\mathcal{G}_j),
\end{equation}
where $x_i$ denotes sequence-derived information of $p_i$, $\mathcal{G}_i$ denotes its biomedical graph context, and $\Theta$ denotes trainable parameters.

\textbf{Protein views.}
Each protein is represented from two complementary views. The sequence view is an ordered set of region-level inputs $x_i=\{x_{i1},\ldots,x_{iL_i}\}$ obtained from amino-acid segments. The graph view is built from $M$ protein-entity bipartite graphs $\mathcal{G}^{m}=(\mathcal{P},\mathcal{E}^{m},\mathcal{R}^{m})$, where $\mathcal{E}^{m}$ is a modality-specific entity set and $\mathcal{R}^{m}$ records observed protein-entity links. In this work, $M=4$ and the modalities correspond to drugs, diseases, miRNAs, and lncRNAs.

\textbf{Learning objective.}
The central difficulty is information imbalance: a novel protein may have no PPI links, incomplete biomedical associations, or uneven coverage across graph modalities. \method\ therefore learns one sequence embedding and multiple graph embeddings for each protein, then estimates a pair label by adaptively weighting these views. The model is trained with binary PPI supervision and an auxiliary graph reconstruction signal that encourages proteins sharing biomedical entities to occupy compatible graph neighborhoods.

\section{Proposed Method}
\begin{figure}[t]
\centering
\includegraphics[width=\textwidth]{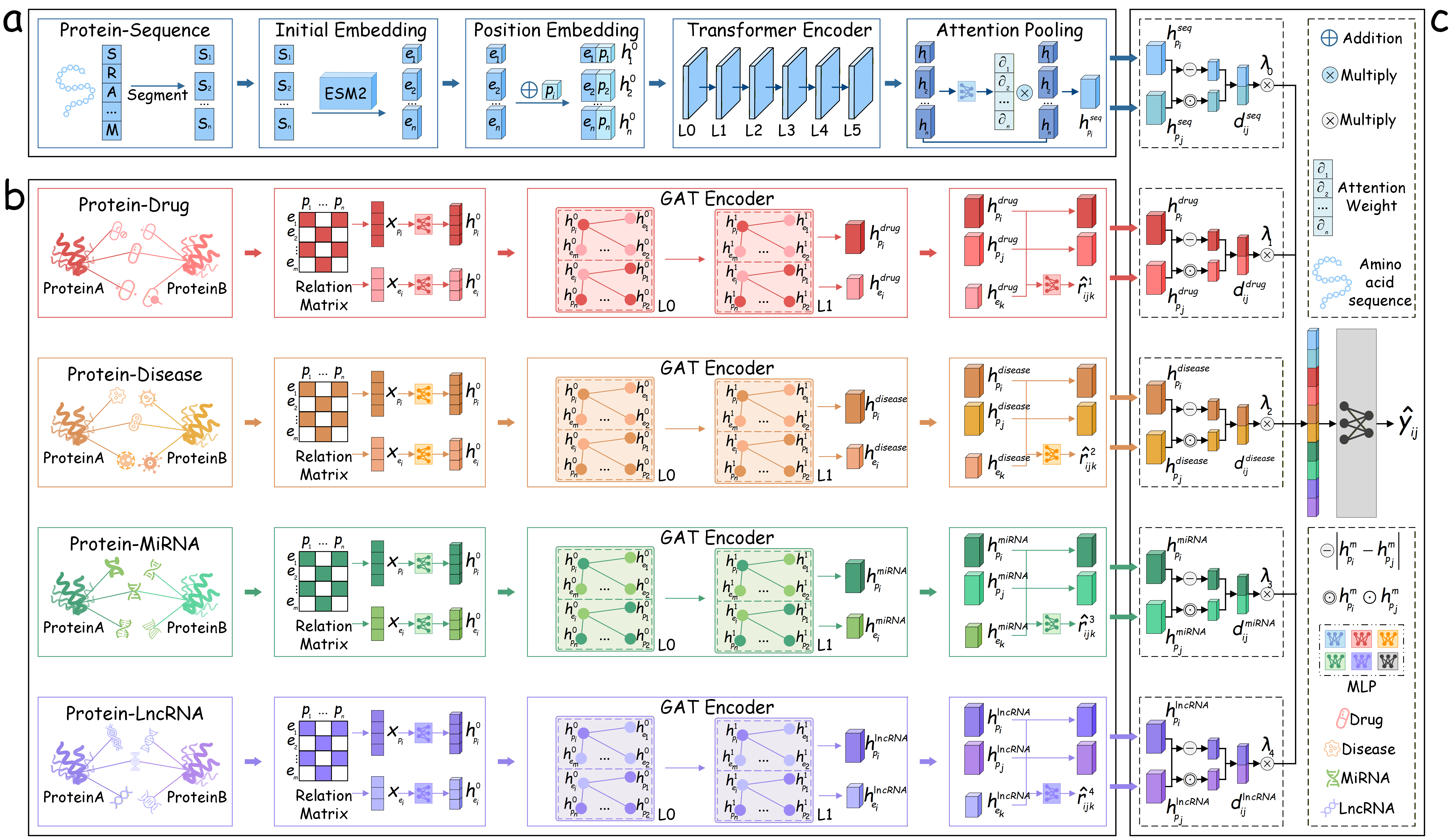}
\caption{Overview of \method. The model combines a region-aware protein sequence branch, multimodal knowledge-graph branches, bridge reconstruction, and pair-level gated fusion for cold-start PPI prediction.}
\label{fig:model_structure}
\end{figure}

\textbf{Overview.}
Figure~\ref{fig:model_structure} summarizes \method. The framework first obtains a sequence representation from region-level protein language model features and obtains graph representations from modality-specific protein-entity graphs. It then regularizes graph embeddings with bridge reconstruction and predicts PPI labels through pair-level gated fusion.

\subsection{Sequence and Graph Encoding}
For the sequence view, region features generated by a frozen ESM2 encoder are passed to a Transformer encoder. Given contextual region vectors $\{u_{i1},\ldots,u_{iL_i}\}$, \method\ forms a protein sequence embedding by soft region aggregation:
\begin{equation}
  s_i=\sum_{t=1}^{L_i}\alpha_{it}u_{it}, \quad
  \alpha_{it}=\operatorname{softmax}_{t}\left(q_s^\top \tanh(W_su_{it})\right).
\end{equation}
This aggregation compresses variable-length region sequences into a fixed-dimensional representation while retaining region-level context.

For the graph view, each modality graph $\mathcal{G}^{m}$ is encoded independently. Let $h_u^{m,\ell}$ be the representation of node $u$ at layer $\ell$. A graph attention update can be written compactly as
\begin{equation}
  h_u^{m,\ell+1}
  =
  \sigma\left(
  \sum_{v\in\mathcal{N}_m(u)}
  a_{uv}^{m,\ell} W_m^\ell h_v^{m,\ell}
  \right),
  \quad
  a_{uv}^{m,\ell}=\operatorname{softmax}_{v}\left(\rho_m^\top[W_m^\ell h_u^{m,\ell}\Vert W_m^\ell h_v^{m,\ell}]\right).
\end{equation}
After the final layer, the modality-specific protein representation is denoted by $g_i^m$.

\subsection{Bridge-Regularized Pair Learning}
Biomedical graph links are sparse, especially for proteins appearing in cold-start evaluation. To regularize graph representations, \method\ reconstructs whether two proteins share an entity in a modality graph. For sampled triples $(p_i,p_j,e_k)$, define
\begin{equation}
  r_{ijk}^{m}=I[e_k\in\mathcal{N}_m(p_i)\cap\mathcal{N}_m(p_j)], \quad
  \hat{r}_{ijk}^{m}=\operatorname{MLP}_m([g_i^m\Vert g_j^m\Vert h_{e_k}^m]).
\end{equation}
This objective encourages proteins connected to similar biomedical contexts to obtain compatible modality embeddings, providing a transfer signal when direct PPI observations are absent.

\subsection{Adaptive Fusion and Objective}
For each branch $b\in\{seq,1,\ldots,M\}$, \method\ constructs a symmetric pair vector
\begin{equation}
  d_{ij}^{b}=[|h_i^b-h_j^b|\Vert h_i^b\odot h_j^b],
\end{equation}
where $h_i^{seq}=s_i$ and $h_i^m=g_i^m$ for graph modality $m$. A gate then assigns pair-specific branch weights and produces the final predictor:
\begin{equation}
  \lambda_{ij}=\operatorname{softmax}(W_g[d_{ij}^{seq}\Vert d_{ij}^{1}\Vert\cdots\Vert d_{ij}^{M}]), \quad
  \hat{y}_{ij}=\sigma\left(W_o[\lambda_{ij}^{seq}d_{ij}^{seq}\Vert\lambda_{ij}^{1}d_{ij}^{1}\Vert\cdots\Vert\lambda_{ij}^{M}d_{ij}^{M}]\right).
\end{equation}
The complete training criterion is
\begin{equation}
  \mathcal{L}
  =
  \frac{1}{|\mathcal{D}|}\sum_{(i,j)\in\mathcal{D}}\bce(\hat{y}_{ij},y_{ij})
  +
  \eta\sum_{m=1}^{M}
  \frac{1}{|\mathcal{S}^m|}\sum_{(i,j,k)\in\mathcal{S}^m}\bce(\hat{r}_{ijk}^{m},r_{ijk}^{m}),
\end{equation}
where $\mathcal{D}$ is the supervised PPI training set, $\mathcal{S}^m$ is the sampled bridge set of modality $m$, and $\eta$ balances PPI supervision and bridge reconstruction.

\section{Experiments}
\subsection{Experimental Settings}
\textbf{Datasets.}
We evaluate \method\ on two PPI benchmark datasets with multimodal protein knowledge graphs. Dataset1 is derived from MTV-PPI, and Dataset2 is constructed from public biological databases including STRING, DrugBank, LncTarD, miRTarBase, and CTD. Both datasets contain protein sequences and four types of protein-associated biomedical entities: drugs, diseases, miRNAs, and lncRNAs. For each dataset, proteins are split at protein level into old and novel groups. Old-old pairs are used for training, with 10\% held out for validation, while novel-old and novel-novel pairs are used for cold-start testing.

\textbf{Metrics and baselines.}
We report Accuracy (ACC), Sensitivity (SEN), Precision (PRE), F1-score (F1), ROC-AUC (AUC), PR-AUC (AUPR), and Matthews Correlation Coefficient (MCC). \method\ is compared with TAGPPI \cite{TAGPPI}, HNSPPI \cite{HNSPPI}, EResCNN \cite{EResCNN}, BaPPI \cite{BaPPI}, KGF-GNN \cite{KGF}, HEENN \cite{HEENN}, and ESM2\_AMP \cite{ESM2_AMP}. Training uses 100 epochs, batch size 1024, learning rate $6\times 10^{-4}$, and 256-dimensional embeddings. The sequence and graph encoders use a 6-layer Transformer and a 2-layer GAT, respectively.

\subsection{Comparative Performance}
\begin{table}[t]
\centering
\scriptsize
\setlength{\tabcolsep}{2pt}
\renewcommand{\arraystretch}{0.95}
\resizebox{\textwidth}{!}{
\begin{tabular}{cccccccccc}
\toprule
Dataset & Metric & TAGPPI & HNSPPI & EResCNN & BaPPI & KGF & HEENN & ESM2-AMP & \method \\
\midrule
\multirow{7}{*}{Dataset1}
& Acc  & $59.39{\pm}1.18$ & $63.38{\pm}0.41$ & $65.38{\pm}1.76$ & $66.40{\pm}1.41$ & $56.71{\pm}4.38$ & $62.81{\pm}1.46$ & $77.99{\pm}0.81$ & $\mathbf{81.31{\pm}1.28}$ \\
& Sen  & $38.17{\pm}3.76$ & $63.72{\pm}0.91$ & $44.13{\pm}5.74$ & $73.06{\pm}1.62$ & $22.54{\pm}2.90$ & $51.83{\pm}25.53$ & $77.02{\pm}2.57$ & $\mathbf{78.41{\pm}3.10}$ \\
& Pre  & $66.80{\pm}3.01$ & $63.56{\pm}0.78$ & $77.14{\pm}2.17$ & $64.78{\pm}2.69$ & $79.52{\pm}0.49$ & $70.50{\pm}7.64$ & $78.67{\pm}1.10$ & $\mathbf{83.61{\pm}1.40}$ \\
& F1   & $48.42{\pm}2.59$ & $63.39{\pm}0.40$ & $55.92{\pm}4.32$ & $68.62{\pm}1.06$ & $35.08{\pm}3.57$ & $56.18{\pm}8.95$ & $77.80{\pm}0.96$ & $\mathbf{80.89{\pm}1.40}$ \\
& Auc  & $62.60{\pm}2.03$ & $68.23{\pm}0.79$ & $75.39{\pm}1.24$ & $71.96{\pm}0.71$ & $75.16{\pm}2.81$ & $73.98{\pm}1.71$ & $85.68{\pm}0.71$ & $\mathbf{88.79{\pm}0.98}$ \\
& Aupr & $65.11{\pm}1.35$ & $68.23{\pm}0.63$ & $74.77{\pm}1.14$ & $69.08{\pm}1.69$ & $72.39{\pm}2.06$ & $71.12{\pm}1.00$ & $84.90{\pm}0.75$ & $\mathbf{88.38{\pm}0.54}$ \\
& Mcc  & $21.02{\pm}2.36$ & $27.24{\pm}0.81$ & $34.18{\pm}2.66$ & $33.12{\pm}2.39$ & $24.51{\pm}1.96$ & $30.43{\pm}3.78$ & $56.04{\pm}1.57$ & $\mathbf{62.83{\pm}2.35}$ \\
\midrule
\multirow{7}{*}{Dataset2}
& Acc  & $68.52{\pm}1.37$ & $57.20{\pm}0.50$ & $55.38{\pm}0.47$ & $62.95{\pm}3.18$ & $62.80{\pm}0.79$ & $66.33{\pm}3.67$ & $58.72{\pm}0.86$ & $\mathbf{84.61{\pm}0.53}$ \\
& Sen  & $57.90{\pm}5.71$ & $57.18{\pm}0.77$ & $\mathbf{84.91{\pm}0.64}$ & $62.95{\pm}2.88$ & $42.88{\pm}3.42$ & $48.72{\pm}12.75$ & $67.97{\pm}10.56$ & $77.44{\pm}0.96$ \\
& Pre  & $73.83{\pm}1.96$ & $57.23{\pm}0.65$ & $53.41{\pm}0.52$ & $46.10{\pm}3.11$ & $71.39{\pm}1.19$ & $76.36{\pm}6.03$ & $57.57{\pm}2.38$ & $\mathbf{90.43{\pm}0.54}$ \\
& F1   & $64.66{\pm}2.95$ & $57.20{\pm}0.50$ & $65.57{\pm}0.33$ & $53.18{\pm}2.58$ & $53.49{\pm}2.51$ & $58.34{\pm}7.99$ & $61.95{\pm}2.93$ & $\mathbf{83.43{\pm}0.51}$ \\
& Auc  & $77.12{\pm}1.11$ & $61.39{\pm}0.79$ & $55.34{\pm}0.50$ & $68.67{\pm}2.00$ & $69.88{\pm}0.87$ & $75.24{\pm}3.50$ & $62.71{\pm}0.76$ & $\mathbf{92.27{\pm}0.45}$ \\
& Aupr & $75.42{\pm}1.07$ & $63.42{\pm}0.51$ & $53.19{\pm}0.56$ & $51.78{\pm}0.90$ & $69.01{\pm}0.86$ & $75.41{\pm}3.68$ & $64.26{\pm}0.50$ & $\mathbf{92.72{\pm}0.25}$ \\
& Mcc  & $38.15{\pm}2.04$ & $14.40{\pm}1.01$ & $13.30{\pm}1.06$ & $24.56{\pm}5.34$ & $27.97{\pm}1.11$ & $35.34{\pm}6.42$ & $18.27{\pm}0.47$ & $\mathbf{69.95{\pm}0.95}$ \\
\bottomrule
\end{tabular}}
\caption{Performance comparison on Task 1, where test pairs contain one novel protein and one old protein.}
\label{tab:comparison_task1}
\end{table}

\begin{table}[t]
\centering
\scriptsize
\setlength{\tabcolsep}{2pt}
\renewcommand{\arraystretch}{0.95}
\resizebox{\textwidth}{!}{
\begin{tabular}{cccccccccc}
\toprule
Dataset & Metric & TAGPPI & HNSPPI & EResCNN & BaPPI & KGF & HEENN & ESM2-AMP & \method \\
\midrule
\multirow{7}{*}{Dataset1}
& Acc  & $54.73{\pm}2.56$ & $50.22{\pm}1.31$ & $52.75{\pm}1.89$ & $57.48{\pm}2.28$ & $53.33{\pm}3.63$ & $53.41{\pm}3.35$ & $59.23{\pm}1.60$ & $\mathbf{74.40{\pm}1.85}$ \\
& Sen  & $46.04{\pm}18.44$ & $50.25{\pm}1.87$ & $8.18{\pm}1.61$ & $59.91{\pm}6.33$ & $16.48{\pm}15.22$ & $18.36{\pm}6.86$ & $\mathbf{79.36{\pm}9.69}$ & $68.47{\pm}7.83$ \\
& Pre  & $56.94{\pm}7.35$ & $49.73{\pm}2.63$ & $\mathbf{79.16{\pm}5.12}$ & $57.24{\pm}5.00$ & $53.97{\pm}5.44$ & $60.46{\pm}5.24$ & $56.24{\pm}2.98$ & $77.43{\pm}3.79$ \\
& F1   & $49.47{\pm}7.58$ & $49.92{\pm}1.05$ & $14.78{\pm}2.69$ & $58.22{\pm}3.12$ & $22.86{\pm}17.60$ & $27.46{\pm}7.96$ & $65.59{\pm}0.36$ & $\mathbf{72.34{\pm}3.75}$ \\
& Auc  & $57.60{\pm}3.31$ & $50.53{\pm}1.91$ & $63.56{\pm}2.54$ & $59.23{\pm}2.73$ & $52.59{\pm}5.20$ & $56.07{\pm}2.85$ & $64.14{\pm}2.55$ & $\mathbf{82.15{\pm}2.75}$ \\
& Aupr & $57.43{\pm}5.17$ & $50.07{\pm}3.75$ & $63.82{\pm}2.87$ & $52.24{\pm}10.26$ & $49.28{\pm}1.10$ & $54.51{\pm}1.45$ & $59.47{\pm}13.40$ & $\mathbf{82.00{\pm}2.61}$ \\
& Mcc  & $10.90{\pm}4.93$ & $0.53{\pm}2.60$ & $13.54{\pm}1.89$ & $14.56{\pm}4.74$ & $2.86{\pm}3.92$ & $8.36{\pm}4.14$ & $20.44{\pm}2.82$ & $\mathbf{48.85{\pm}3.59}$ \\
\midrule
\multirow{7}{*}{Dataset2}
& Acc  & $59.05{\pm}3.03$ & $51.47{\pm}0.54$ & $54.84{\pm}0.90$ & $59.29{\pm}2.15$ & $53.39{\pm}1.03$ & $55.79{\pm}0.71$ & $58.29{\pm}2.00$ & $\mathbf{79.56{\pm}1.85}$ \\
& Sen  & $53.77{\pm}8.78$ & $51.48{\pm}0.69$ & $\mathbf{83.79{\pm}2.34}$ & $51.67{\pm}2.46$ & $49.47{\pm}12.44$ & $15.11{\pm}4.90$ & $66.76{\pm}9.90$ & $67.53{\pm}4.01$ \\
& Pre  & $60.54{\pm}2.91$ & $51.28{\pm}1.78$ & $52.96{\pm}1.39$ & $40.11{\pm}0.86$ & $52.86{\pm}2.44$ & $86.21{\pm}4.94$ & $55.53{\pm}3.18$ & $\mathbf{88.78{\pm}1.11}$ \\
& F1   & $56.62{\pm}5.46$ & $51.37{\pm}0.95$ & $64.87{\pm}0.83$ & $45.15{\pm}1.44$ & $50.74{\pm}7.84$ & $25.40{\pm}6.84$ & $60.13{\pm}2.29$ & $\mathbf{76.65{\pm}2.42}$ \\
& Auc  & $62.74{\pm}4.74$ & $52.66{\pm}0.67$ & $54.69{\pm}0.83$ & $58.14{\pm}1.72$ & $53.28{\pm}1.95$ & $69.29{\pm}1.34$ & $60.82{\pm}1.82$ & $\mathbf{89.09{\pm}1.57}$ \\
& Aupr & $61.31{\pm}4.07$ & $52.42{\pm}1.14$ & $52.89{\pm}1.31$ & $36.74{\pm}4.08$ & $52.19{\pm}3.42$ & $70.66{\pm}1.24$ & $54.76{\pm}10.08$ & $\mathbf{89.15{\pm}1.34}$ \\
& Mcc  & $18.35{\pm}5.87$ & $2.97{\pm}1.09$ & $12.19{\pm}2.44$ & $13.84{\pm}3.47$ & $6.21{\pm}2.29$ & $21.74{\pm}1.92$ & $15.19{\pm}2.64$ & $\mathbf{60.90{\pm}2.74}$ \\
\bottomrule
\end{tabular}}
\caption{Performance comparison on Task 2, where both proteins in each test pair are novel proteins.}
\label{tab:comparison_task2}
\end{table}

Tables~\ref{tab:comparison_task1} and~\ref{tab:comparison_task2} show that \method\ achieves the strongest overall performance in both cold-start settings. On Task 1, \method\ obtains the best score for all metrics on Dataset1 and for six of seven metrics on Dataset2. The only exception is SEN on Dataset2, where EResCNN reaches higher recall but has substantially lower PRE, F1, AUC, AUPR, and MCC. This pattern suggests that \method\ provides a more balanced decision boundary rather than only increasing positive predictions.

Task 2 is more difficult because neither protein in the candidate pair has observed PPI edges during training. \method\ still obtains the best ACC, F1, AUC, AUPR, and MCC on both datasets. The gains are especially clear for AUC, AUPR, and MCC, which are sensitive to ranking quality and balanced binary prediction. These results support the central design of \method: sequence information supplies intrinsic protein features, biomedical knowledge graphs provide external relational context, and gated fusion adjusts the use of each modality for the candidate pair.

\section{Conclusion}
This paper presented \method, a multimodal sequence-graph framework for cold-start PPI prediction. \method\ combines region-aware sequence representation, modality-specific graph attention encoders, bridge reconstruction, and pair-level gated fusion. Experiments on two datasets under novel-old and novel-novel cold-start settings show consistent improvements over representative PPI prediction baselines. Future work will extend \method\ to additional biomedical modalities and evaluate its transferability across species and disease-specific interaction networks.

\bibliographystyle{unsrt}
\bibliography{reference}

\end{document}